\documentclass[11pt]{article}
\usepackage{eamt22}
\usepackage{times}
\usepackage{url}
\usepackage{latexsym}
\usepackage{amsfonts}
\usepackage[small,bf]{caption} 
\newcommand{\jabert}{https://huggingface.co/cl-tohoku/bert-base-japanese}
\newcommand{\corpora}{http://www.phontron.com/japanese-translation-data.php}

\newcommand{\bftab}{\fontseries{b}\selectfont}

\title{Learning a Formality-Aware Japanese Sentence Representation}

\author{Henry Li Xinyuan \and Ray Lee \and Jerry Chen \and Kelly Marchisio \\ Johns Hopkins University \\ USA}

\date{}

\begin{document}
\maketitle
\begin{abstract}

While the way intermediate representations are generated in encoder-decoder sequence-to-sequence models typically allow them to preserve the semantics of the input sentence, input features such as formality might be left out. On the other hand, downstream tasks such as translation would benefit from working with a sentence representation that preserves formality in addition to semantics, so as to generate sentences with the appropriate level of social formality -- the difference between speaking to a friend versus speaking with a supervisor. We propose a sequence-to-sequence method for learning a formality-aware representation for Japanese sentences, where sentence generation is conditioned on both the original representation of the input sentence, and a side constraint which guides the sentence representation towards preserving formality information. Additionally, we propose augmenting the sentence representation with a learned representation of formality which facilitates the extraction of formality in downstream tasks. We address the lack of formality-annotated parallel data by adapting previous works on procedural formality classification of Japanese sentences. Experimental results suggest that our techniques not only helps the decoder recover the formality of the input sentence, but also slightly improves the preservation of input sentence semantics.

\end{abstract}

\section{Introduction}

\subsection{Motivation}

Choosing appropriate levels of formality depending on ones audience is almost universal across languages \cite{Heylighen:99}.  Whereas a speaker might use slang words and casual word choice when conversing with friends, many people use more formal language when speaking to a supervisor, for instance. This is just one example of downstream tasks where generating sentences with the correct formality is crucial. While general sequence-to-sequence models are now capable of producing results that are very coherent and semantically close to the target output, details such as formality can often be neglected by the model \cite{rao-tetreault-2018-dear} - details that can make or break a career if approached carelessly by a human. As such, we would like to let our models learn formality-aware sentence representations that would allow both formality and semantics to be recovered for downstream tasks.

\subsection{Related Studies}

\subsubsection*{Formality Transfer}

A closely related and comparatively more well-studied problem to our proposed study of formality-aware sentence representations is that of formality transfer, where one seeks to build a model that takes a sentence and convert it to a semantically similar sentence in a different formality class \cite{chawla-yang-2020-semi}. An important distinction between our study and that of formality transfer is that we don't attempt to change the formality of input sentences - such a study would require parallel sentences in different formality classes \cite{rao-tetreault-2018-dear}, which we don't have access to. Just like in formality transfer studies, we evaluate our model by the percentage of sentences that are in the correct formality class (in our case, the correct formality class is the formality class of the input sentence).

Formality transfer can be considered a special case of the more general problem of text style transfer, a task where one seeks to control the style of generated text while preserving content \cite{jin2022survey}. Many of the recent studies on text style transfer with parallel data have made use of the encoder-decoder model architecture, building on top of popular models such as LSTM \cite{rao-tetreault-2018-dear,niu-etal-2018-multi}. More recently, due to the rise in popularity of the Transformer architecture \cite{vaswani2017attention} for machine translation and other text generation tasks, studies on text style transfer have shifted towards using Transformer as well, many of which \cite{DBLP:journals/corr/abs-2104-05196,liu-etal-2021-learning,NEURIPS2020_7a677bb4} do so by extending pre-trained Transformer-based language models such as GPT2 \cite{Radford2019LanguageMA}. Due to there not being as large and powerful a pre-trained Japanese language model as GPT2, as well as our need to incorporate machine translation as one of our joint tasks, we use Transformer as the basis of our formality transfer model.

One of the earliest and most influential studies on style transfer was done by Sennrich et al.~\shortcite{sennrich-etal-2016-controlling}, who showed that appending a style tag on the source sentence can improve the quality of style transfer as the model learns to attend to the formality tag. A large number of subsequent studies on style transfer, up to this day, are fundamentally based upon this idea of style tag augmentation to input sentences. This setup has also been applied to formality transfer by Niu et al.~\shortcite{niu-etal-2018-multi} with some success. 

Since then, much of the focus has been shifted to proposing modifications the base encoder-decoder architecture in order to further improve style transfer. 
Fu et al.~\shortcite{Fu_Tan_Peng_Zhao_Yan_2018}, and later Marchisio et al.\shortcite{marchisio-etal-2019-controlling}, proposed a double-decoder architecture where each decoder is responsible for a different formality class. Liu et al.~\shortcite{liu-etal-2021-learning} focused on augmenting the objective, rewarding the model not only for the accuracy and quality of the generated sentences, but also for their style correctness. Similarly, Wang et al.~\shortcite{Wang2020FormalityST} designed a joint objective between word sequence cross-entropy loss and an additional loss term which penalises generated sentences that have the incorrect formality. Such an approach would require on-the-fly classification/verification of style, which can be addressed by training a style classifier prior to training the style transfer generator. 

Our proposed model is fundamentally a combination of these three techniques. Inspired by the style-embedding model proposed by Fu et al.~\shortcite{Fu_Tan_Peng_Zhao_Yan_2018}, we condition the sentence generation process on both the input sentence representation and the desired style. Instead of prepending the input sentences with a formality tag or training separate decoders for different formality classes, we propose instead the explicit extension of the encoder with a learnable formality mask, applied on the encoder output additively or multiplicatively. By using a joint objective between sentence reconstruction quality and formality correctness of output sentences, we make sure that the generated sentences---and by extension the formality-augmented encoding---preserve both semantics and formality of the input sentences.

\subsubsection*{Formality Across Languages}

\begin{figure}
  \includegraphics[scale=.5]{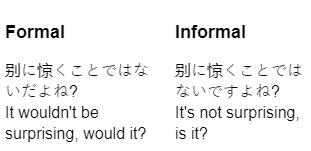}
  \caption{Formal vs. informal sentences in Japanese and in English}\label{fig:sentences}
\end{figure}

Formality might be common across languages, yet there are certain languages that have more clearly identifiable formality markers. English, despite being the most well-studied language for in the context of formality style transfer \cite{jin2022survey}, lacks a simple morphology-based formality marker, rendering its formality classification a non-trivial problem \cite{Sheikha2012LearningTC,5587767}. Past studies have therefore had to rely on lexical or contextual cues for formality classification \cite{pavlick-tetreault-2016-empirical}, or rule-based formality annotation based on parse trees \cite{DBLP:journals/corr/abs-2104-05196}. By contrast, while no such formality-annotated corpus as the GYAFC dataset \cite{rao-tetreault-2018-dear} exist for Japanese, formality in Japanese can be deduced directly from its verb conjugation. In addition, the vast majority of formality transfer studies have focused on English, with a number of others on languages such as French, Italian, Breton and Portuguese \cite{briakou-etal-2021-ola}, as well as on Estonian and Latvian \cite{DBLP:journals/corr/abs-1903-11283}. Past works on Japanese formality transfer in neural systems are very limited. We therefore focus our study on the relatively accessible and less well-studied problem of formality transfer for Japanese. We address the lack of parallel data by adapting a previous rule-based Japanese formality converter by Feely et al.~\shortcite{Feely:19}, similar to the automatic annotation proposed by Sennrich et al.~\shortcite{sennrich-etal-2016-controlling}. We discuss the construction of our parallel data in detail in Section $3$.

\section{Learning a Formality-Aware Sentence Representation}

\subsection{Extracting an Intermediate Representation from a Base Sentence Autoencoder}

Let $g_0: X \rightarrow X$, which maps the set of all possible Japanese sentences $X$ onto itself through an intermediate vector representation, be our Japanese autoencoder. We would like our autoencoder to preserve the semantics of the input sentence as much as possible, therefore for $g_0(x) = \hat{x}$ we train $g_0$ to minimise $-\log (P(\hat{x} | x))$ for $x, \hat{x} \in X$.

\subsubsection*{Machine Translation as the Base Task}

We consider an alternative base task: that of machine translation. Here we have $g_0: X \rightarrow Y$ which maps the set of all possible sentences in the source language $X$ onto the set of target language sentences $Y$, and $g_0$ is trained to minimise the negative log-likelihood $-\log (P(\hat{y} | x))$ for $x \in X$, $\hat{y} \in Y$ and $g_0(x) = \hat{y}$.

There are a number of issues associated with using machine translation as the base task for learning a formality-aware representation. First is the inherent difficulty in machine translation, especially between language pairs with very different grammatical characteristics as Japanese and English \cite{Matsumura2017EnglishJapaneseNM}: sentences that are not at least meaningfully coherent can't have their formality evaluated. Another issue is that formality may not map across languages - we will discuss this issue in more detail in later sections. 

\subsection{Sentence Formality Classifier}

Let $S := \{0, 1\}$ denote the set of possible formality style of a Japanese sentence. Let $c^*: X \rightarrow S$ denote the oracle formality mapping function. For some sentence $x \in X $, $c^*(x) = 1$ if $x$ is formal and vice-versa. We make the simplifying assumption that exactly $2$ distinct formality classes exist in Japanese. We thus define the task of sentence formality classification as finding some model $c: X \rightarrow S$ which minimises the logistic regression loss $-\log(\prod_{c^*(x) = 1}c(x) \prod_{c^*(x) = 0}(1 - c(x)))$. Performance on the task is measured by the proportion of correctly classified sentences.

\subsubsection*{Sentence Formality Classifier from Pre-trained Encoder}

Let $g_p: X \rightarrow \mathbb{R}^k$ be the encoder of some pre-trained language model with an encoder-decoder architecture: that is, for some sentence $x \in X$, $g_p(x)$ maps $x$ to some vector $T^k$ in the $k$-dimensional real vector space. One such mapping can be constructed by running an encoder forward pass on the input sentence, taking the pooled output of BERT \cite{devlin2019bert} as the output. Given past studies that BERT is able to pick up on phrase-level characteristics of language \cite{jawahar-etal-2019-bert}, we hypothesise that it would be possible to train a classifier to use the output vector $T^k$ to classify the formality class of a sentence. In our subsequent study, we benchmark our results against such a linear classifier trained on the output of a pre-trained Japanese BERT model\footnote{\jabert} \cite{JapaneseBert}.

\subsubsection*{Practicalities of the Formality Classifier}

The rule-based formality classifier unfortunately cannot be back-propagated through the model, as such, we define the formality classifier $c: X \rightarrow S$ as follows: let autoencoder $g:= X \rightarrow X$ be the composition between an encoder $e_{g}: X \rightarrow \mathbb{R}^k$ and a decoder $d_{g}: \mathbb{R}^k \rightarrow X$. We predict the formality of the sentence with a linear classifier $c_e: \mathbb{R}^k \rightarrow S$ which takes the encoder output as its input. Thus we have $c:= c_e \circ e_{g}$ as our back-propagable classifier.

In practice, we observe that our formality classifier is highly accurate (as can be seen in Table $1$), therefore we take our classifier $c$ as an approximation of the oracle $c^*$.

\subsection{Sentence Reconstruction and Formality Classification as a Joint Task}

We propose the combination of the two above-mentioned models, $g_1: X \rightarrow (X, S)$, as the simplest joint-training approach for learning a formality-preserving sentence representation. We construct the objective of our joint task model as a linear combination of the objectives of the two tasks; in other words, for $g_1(x) = (\hat{x}, \hat{s})$, we try to minimise $(1 - \lambda) (-\log (P(\hat{x} | x))) - \lambda \log(\prod_{c^*(x) = 1} s \prod_{c^*(x) = 0}(1 - s))$. We evaluate performance on this task both by the sentence reconstruction quality (measured by BLEU), and by the accuracy of the formality classifier.

\subsection{Sentence Reconstruction with Correctness of Output Sentence Formality as a side constraint}

One of the most well-known studies on formality-controlled sentence generation, by Sennrich et al.~\shortcite{sennrich-etal-2016-controlling}, proposed using the formality of output sentences as a side constraint for the model during training. Given the same sentence reconstruction model $g: X \rightarrow X$ as before with $g_2(x) = \hat{x}$, let $c: Y \rightarrow \{0, 1\}$ be the our formality classifier (which we use to approximate $c^*$). $g_2$ would then be trained to minimise the objective $(1 - \lambda) (-\log (P(\hat{y} | x))) - \lambda \log(\prod_{c^*(x) = 1}(c(\hat{y})) \prod_{c^*(x) = 0}(1 - c(\hat{y})))$: a linear combination between cross entropy loss on words in the sentence, and the Bernoulli cross entropy loss between the expected formality and the actual formality of the output sentence. The formality classifier $c$ should be pre-trained so that it can continue to be effective as the model is trained. 

By letting the side constraint be the formality of output sentences rather than the output of a classifier which works with the input sentence representation, we address the potential over-fitting issues that collapsing a high-dimensional representation into a very low dimensional output (in the case of formality labels, only $2$-dimensional) might cause. We can also demonstrate in this very simple setting that downstream tasks can indeed extract both semantics and formality from the learned representation.

\subsection{Formality-augmented Autoencoder}

We propose a modified autoencoder $g_3: X \rightarrow X$ on top of $g_2$, consisting of an encoder $e_{g_3}: X \rightarrow \mathbb{R}^k$ and a decoder $d_{g_3}: \mathbb{R}^k \rightarrow X$. Instead of simply using a Transformer encoder $e_t$ as $e_{g_3}$, we instead define $e_{g_3}$ as follows:

\begin{itemize}
    \item We define $f: (\mathbb{R}^k, S) \rightarrow \mathbb{R}^k$ as our ``formality augmentation" function: $f$ ``augments" the original encoder output, facilitating the decoder's task of recovering the formality class of the input sentence; 
    \item Given some formality classifier $c$, we define $e_{g_3} := f(e_t(x), c(x))$ for some sentence $x \in X$. 
\end{itemize}

\begin{figure*}
\centering
\includegraphics[]{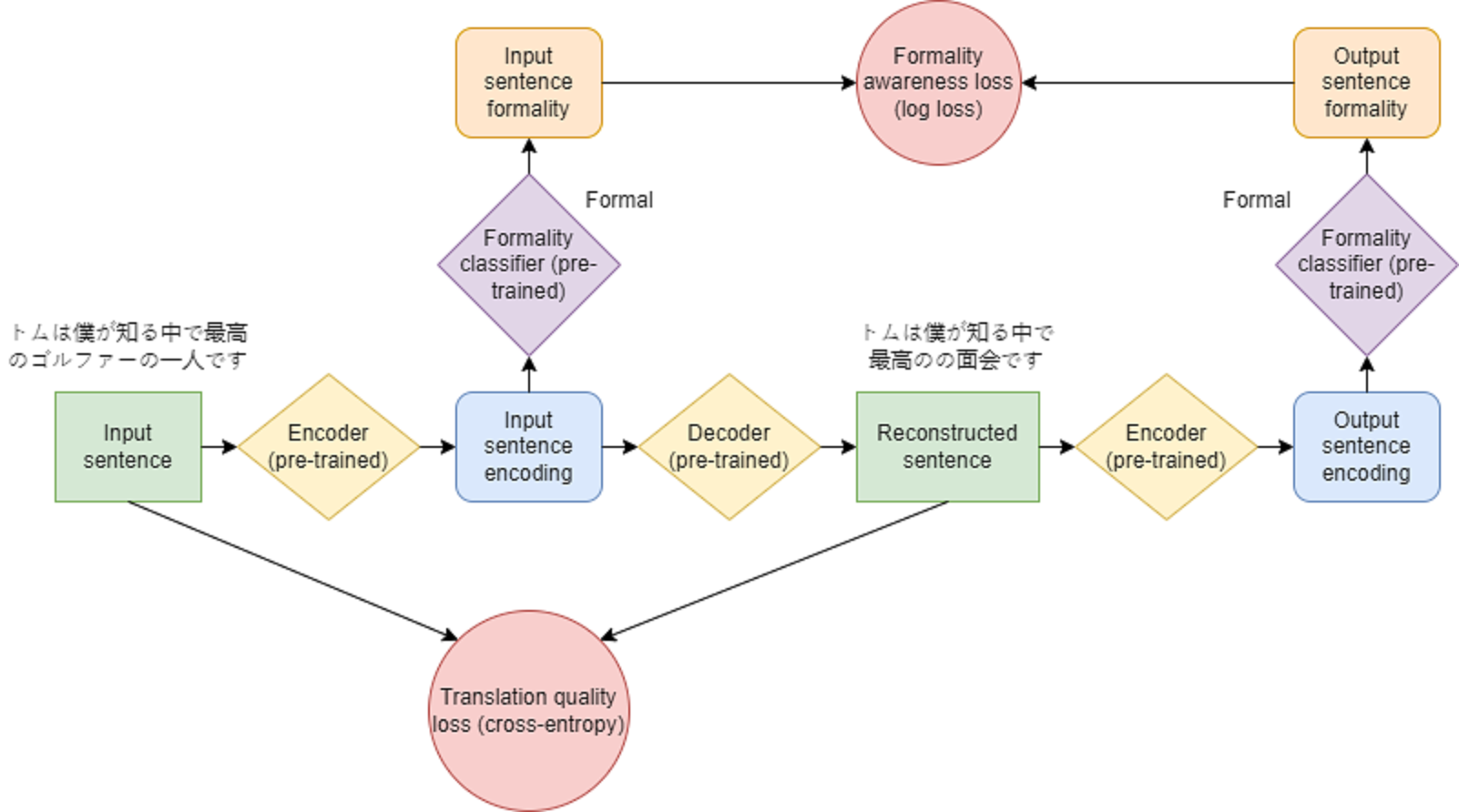}
\caption{Formality-augmented Sentence Reconstruction: Model Architecture}\label{fig:diagram}
\end{figure*}

Unlike $c$, the formality augmentation function $f$ can be trained alongside the rest of the model. The training objective is the same linear combination between translation quality and correctness of the formality of output sentences as the one we described in Section $2.4$. Our model architecture for the formality-augmented sentence reconstruction task is illustrated in Figure~\ref{fig:diagram}.

Let $g_3(x) = x$ and $c(x) = \hat{s}$, our technique can be described mathematically as as follows: $P(\hat{x} | x) = P(\hat{x} | x, \hat{s}) P(\hat{s} | x)$: generating output sentences conditioned on the both the semantics and the formality of the input sentence. As before we use $c$ as an approximation for $c^*$ (and therefore $\hat{s}$ as an approximation for $s = c^*(x)$), and re-write the above formula as $P(\hat{y} | x, \hat{s}) P(\hat{s} | x) \approx P(\hat{y} | x, S) P(s | x)$. Further, as we hypothesised earlier that each sentence is either formal or informal, we have that $P(s | x) = 1$ and therefore $P(\hat{y} | x) \approx P(\hat{y} | x, s) = P(\hat{y} | x)$.

The evaluation of this task likewise is made up of two parts: sentence reconstruction quality, as well as the proper conjugation of the head verb/copula in the output sentence. Since there isn't the need for back-propagation, we use the rule-based formality classifier during our evaluation.

\subsubsection*{Choice of Formality Augmentation Function}

We propose the following two ways of defining the formality augmentation function: 

\begin{itemize}
  \item Additive formality augmentation. The model learns vectors $r_0 \in \mathbb{R}^k$ (informal augmentation vector) and $r_1 \in \mathbb{R}^k$ (formal augmentation vector) and we define $c^*(r, s) = r + r_s$ for $s \in S = \{0, 1\}$ denoting the desired sentence formality;
  \item Multiplicative formality augmentation. We define $r_0$ and $r_1$ similarly as above, and we define $c(r, s) = r \odot r_s$, where $\odot$ denotes element-wise multiplication.
\end{itemize}

The motivation behind an additive formality augmentation is simple: both semantics and style have long been shown to be recoverable with linear vector arithmetic on the embedding space \cite{Ethayarajh2019TowardsUL}. In addition, additive modifications to word embeddings are already used to encode positional information \cite{vaswani2017attention}. While the formality of the input sentence can be recovered from the encoder output (as demonstrated by the excellent performance of our classifier), any formality information would be localised on the embedding for a particular word in the sentence, rather than being present across the length of the sentence embedding vector. As such, any potential misalignment between transformer attention and input words might compromise the decoder's ability to pick up on that formality. The fact that Japanese words usually consist of multiple characters also means that multiple decoding steps have to pick up and preserve the correct formality marker from the encoder output, further complicating matters. A global shift for the encoder output within the sentence embedding space, with its magnitude and direction defined during training, creates a redundancy for the output words that are affected by formality, while introducing very little difficulty for other words to recover from the shift and be generated correctly. 

By comparison, a multiplicative formality augmentation results in the formality-related shift through the embedding space being proportional to the scale of each dimension of the encoder output. We proceed to compare additive formality augmentation with multiplicative formality augmentation experimentally. 

\section{Experimental Setup}

\subsection{Model Architecture}

We use Transformer \cite{vaswani2017attention} as our base model due to its popularity in many of the downstream tasks, extracting the encoder output as our learned sentence representation. In theory our technique can be generalised to any sequence-to-sequence models with an encoder-decoder architecture \cite{NIPS2014_a14ac55a}. We set the number of encoder and decoder layers to $1$, hidden size to $512$, and the number of encoder and decoder attention heads to $8$. Each encoder layer and each decoder layer share the same configurations. During model training we apply a dropout of $.1$. We use the Adam optimiser \cite{kingma2017adam} during model training, keeping a constant learning rate lambda value of $.0005$.

\subsection{Formality-labelled Parallel Corpus}

While parallel corpora between Japanese and English exist, corpora that are suitable for our purpose are few and far between. We would like our training corpus to consist of a diverse set of contexts and topics, in order to prevent the model from relying on those for formality classification. 

\subsubsection*{Generating Formality Labels}

\begin{figure}
  \includegraphics[scale=.5]{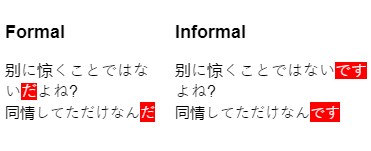}
  \caption{Illustration of rule-based formality classification for Japanese sentences}\label{fig:rule-based}
\end{figure}

We adapted previous work by Feely et al.~\shortcite{Feely:19} which dealt with converting the formality of a Japanese sentence. A Japanese sentence can have at most one head verb/copula; depending on the intended formality of the sentence, the head verb/copula could be conjugated into either honorific form or short form. Any non-head verb or copula are always conjugated into short form (informal). Due to the SOV sentence structure of Japanese, the head verb or copula is always the final verb or copula within a sentence. The formality conversion script by Feely et al. thus achieved its goal by identifying the final verb or copula, classifying it according to the appropriate verb conjugation group, and changing its conjugation according to the desired formality. An example of the rule-based formality classification in progress is shown in Figure~\ref{fig:rule-based}.

We modify the above-mentioned conversion script into a classification script and apply it on the various corpora at our disposal. All legal documents in Japanese are conjugated informally, and our classification script correctly identified all sentences from legal documents as informal. The training dataset that we use does not consist of sentences from legal documents due to concerns about the model using topic to infer formality; instead, we curated a list of relatively simple sentences collected from a number of publicly available Japanese corpora\footnote{\corpora}. 

Some sentences in our curated corpus could not be positively identified by our procedural formality annotation script as either formal or informal due to various reasons, including the sentence not having a head verb or the head verb being too rare. Such sentences were removed from the corpus.

Once our parallel corpus had been annotated with formality, we proceed to split our corpus into separate train/dev/test sets randomly. The training, development and test sets have size $42685$, $5336$ and $5336$, respectively. We have released our dataset at \url{https://doi.org/10.5281/zenodo.6400068}

\subsection{Experiment Settings}

We run the following experiments in order to test the effectiveness of each of the techniques we proposed: 

\begin{itemize}
\item Formality classification with representation generated by pre-trained language model

Here we attempt to extract the formality of a sentence by training a linear classifier on the representation of the sentence generated by the pre-trained Japanese BERT models \cite{JapaneseBert}, as we explained in Section $2.2$.

\item Autoencoder with formality classification as a side constraint

In this experiment, we test the model from Section $2.3$ which would take a Japanese sentence, and attempt to reconstruct the sentence while simultaneously try to produce the correct formality label of the input sentence by using a linear classifier on the sentence representation. We use the previous experiment - direct extraction of formality from pre-trained Japanese BERT - as our baseline. 

\item Sentence reconstruction conditioned on formality

This experiment tests the models described in Sections $2.4$ and $2.5$. Three different settings are tested: no formality augmentation on sentence representation, additive formality augmentation, and multiplicative formality augmentation. Recall that we use $\lambda$ to denote the weight given to the formality correctness objective in the joint objective: we explore the effect of different choices of $\lambda$ to the model's ability to recover both the semantics and the formality of the input sentences.

\item Japanese-English-Japanese back translation conditioned on formality

A cross-lingual experiment is very useful for testing how well our techniques perform on more complicated downstream tasks. On the other hand, since our study is focused on the formality of Japanese, we would like to avoid any having to classify or evaluate the formality transfer of English. As such, we designed the Japanese-English-Japanese circular translation experiment so that our on-the-fly formality classifier can be used during training on the output sentences, and so that our results can be meaningfully evaluated. Cross entropy objective is summed over both the intermediate English sentence and the final target Japanese sentence. 

\end{itemize}

\section{Results}

We first evaluate our formality classifier. Table $1$ shows the results for our Japanese sentence formality classifier, trained as a joint task with a Japanese sentence autoencoder. The strong performance of our classifier gives us the confidence to use the classifier output as an approximation for the actual sentence formality. It also shows that under such a simple joint training setup, it is already possible to preserve formality in the sentence representation.

On the sentence reconstruction and the back translation tasks, we evaluate our model with the following metrics:

\begin{itemize}
  \item Percentage of generated sentences that had the correct formality.
  \item BLEU: Evaluated on the generated sentences against the original sentence. Used to assess semantics preservation of our models.
\end{itemize}

\begin{table*}[]
  \begin{tabular*}{\textwidth}{@{\extracolsep{\fill}} lll}
  \hline
                      & Formality Classification Accuracy \% & Ja-Ja Sentence Reconstruction BLEU \\ \hline
  word-based BERT     & 79.4                                 & -                                  \\
  char-based BERT     & 78.7                                 & -                                  \\
  SR + classification (Ours) & 99.5                                 & 73.0
  \end{tabular*}
  \caption{Formality Classification: word-based BERT and char-based BERT show the baseline results for direct extraction of formality from pre-trained Japanese BERT, with the models in question being based on word-level tokenisation and character-level tokenisation, respectively. SR + classification shows classification accuracy of the formality classifier trained alongside the Japanese sentence autoencoder. BLEU score is evaluated on reconstructed sentences against the corresponding original sentences.}
\end{table*}

\begin{table*}[]
  \begin{tabular*}{\textwidth}{@{\extracolsep{\fill}} lll}
  \hline
                              & \% of Sentences with Correct Formality        & Ja-Ja Sentence Reconstruction BLEU \\ \hline
  No formality augmentation   & 75.4                          & 74.6                               \\
  Multiplicative augmentation & 80.4                          & 70.0                               \\
  Additive augmentation       & \bftab 83.0                   & 77.6
  \end{tabular*}
  \caption{Results for sentence reconstruction conditioned on formality with different formality augmentation techniques. The percentage of output sentences with the correct formality is measured by our rule-based formality classifier, rather than the on-the-fly classifier used during training. BLEU score is evaluated on reconstructed sentences against the corresponding original sentences. The 3 different formality augmentation techniques correspond to the ones discussed in Section 2.5.}
\end{table*}

Table $2$ compares the different techniques for augmenting the encoder output, showing their effects on both formality awareness and the quality of reconstructed sentences. We observe that the highest percentage of reconstructed sentences with the correct formality and the best sentence reconstruction quality are both achieved with additive formality augmentation. This appears to back up the linearity assumption about the word and sentence embedding space. By contrast, while multiplicative formality augmentation is able to improve the formality correctness of generated sentences, this came at the cost of sacrificing sentence reconstruction quality.

\begin{table*}[]
  \begin{tabular*}{\textwidth}{@{\extracolsep{\fill}} lll}
  \hline
  Weight on formality awareness objective  & \% Correct Formality & Ja-Ja Sentence Reconstruction BLEU \\ \hline
  .99 & 81.8                   & 76.3 \\
  .9  & \bftab 83.0            & \bftab 77.6 \\
  .75 & 80.1                   & 73.9 \\
  .5  & 80.6                   & 74.8 \\
  .1  & 79.0                   & 75.7 \\
  .01 & 78.8                   & 76.4
  \end{tabular*}
  \caption{Sentence reconstruction with additive formality augmentation. BLEU is measured on the reconstructed sentences against the original ones. Recall that the cross-entropy objective used to measure sentence reconstruction quality is combined additively with the log loss side constraint objective used to ensure formality correctness of output sentences (the ``formality awareness" objective). Results with different weights for the formality awareness objective are shown.}
\end{table*}

Table $3$ shows more results from the sentence reconstruction experiment, comparing the effect of putting different weights on log loss side constraint which punishes the model for generating sentences with the incorrect formality (the "formality awareness" objective). We observe that putting a high weight on the formality awareness objective not only improves the formality awareness of the sentence reconstruction model, but also improves general reconstructed sentence quality as well. This seems counter-intuitive at first. While we don't have a definitive answer for this phenomenon, we relate this observation to a recent study on the different stages of machine translation model training, namely language modeling, lexical translation and alignment \cite{5bf89fdf23fc4eaeb6b4a8364f22a862}. Before the model training completes the translation (in our case, sentence reconstruction) quality phase, the per-word cross entropy objective dominates even for very large weights on the formality awareness objective. After the translation quality stage, the formality awareness objective dominates the joint objective and shapes the model to generate appropriate honorific or short-form verb suffixes. 

\begin{table*}[]
  \begin{tabular*}{\textwidth}{@{\extracolsep{\fill}} llll}
  \hline
  Augmentation & Formality objective weight & Formality Awareness \% & Ja-En-Ja back translation BLEU \\ \hline
  None  & /   & 26.9                   & \bftab 19.1 \\ 
  Joint & .5  & 38.2                   & 18.2 \\
  Mul   & .9  & 33.7                   & 12.8 \\
  Mul   & .75 & \bftab 67.2            & 13.6 \\
  Mul   & .5  & 64.4                   & 14.6 \\
  Mul   & .1  & 1.7                    & 10.3 \\
  Add   & .5  & 25.8                   & 15.8 
  \end{tabular*}
  \caption{Formality-aware back-translation with additive formality augmentation. The first column shows the formality augmentation used in each experiment: ``Mul" for multiplicative augmentation, ``Add" for additive augmentation, ``Joint" indicating a joint objective with formality without augmentation, and ``None" being the baseline. BLEU is measured on the output sentences against the original sentences. Results with different weights on the formality awareness objective are shown.}
\end{table*}

Table $4$ shows the results for back-translation from Japanese through English. We observe that multiplicative formality augmentation produced strong performance with respect to the baseline performance of the side-constraint training setting, whereas additive augmentation is unable to beat the baseline. On the other hand, unlike in the much simpler sentence reconstruction task, we observe that we are unfortunately forced into a tradeoff between formality awareness and translation quality. This seems to suggest that the formality classes for sentences in our dataset cannot be mapped onto the corresponding English sentences.

\section{Conclusion}

We performed one of the first studies on learning formality-aware representations of Japanese using sequence-to-sequence models. We constructed our own annotated parallel corpus by applying past linguistic studies on Japanese sentence formality, which we made public in order to facilitate future studies on Japanese formality transfer. The rule-based formality annotation based on the conversion script by Feely et al. \shortcite{Feely:19} would benefit from a comparison with other methods of extracting formality, such as those based on a number of linguistic features which was used to construct the GYAFC dataset \cite{pavlick-tetreault-2016-empirical,rao-tetreault-2018-dear}, or unsupervised methods based on contextual evidence \cite{DBLP:journals/corr/abs-1709-02349}. 

We designed a number of different techniques in order to let our model learn a formality-aware representation of Japanese: training a formality classifier along with sentence reconstruction as a joint task, using the formality correctness of output sentences as a side constraint, as well as augmenting sentence representations additively or multiplicatively. Our experiments showed that training a formality classifier along with sentence reconstruction as a joint task allows formality information to be preserved in sentence representations; moreover, the output formality correctness side constraint experiment showed that it is possible to extract both semantics and formality from the formality-preserving sentence representations in downstream tasks. Finally, results suggest that augmenting sentence representations with a learned formality vector allows for slightly improved recovery of formality, and in some cases slightly better preservation of semantics.

We observed that unlike in the case of the simpler sentence reconstruction task, multiplicative formality augmentation is more effective at improving formality transfer in a Japanese-English-Japanese back translation setting. Following up on our brief discussion on the relationship between different methods of embedding augmentation, a study on extracting formality---or style more generally---from a word embedding space would be very helpful in providing insights into the appropriateness of our choice of embedding augmentation methods.

Both the corpus annotation process and the subsequent formality augmentation mechanism introduce minimal overhead to existing sequence-to-sequence models that generate Japanese sentences. The formality augmentation techniques we proposed can easily be adapted to other datasets or languages, while also being relatively easy to generalise into other multi-class sentence generation tasks. Applying our techniques on languages such as English with more available formality-annotated data, on related tasks such as controlling reading level, or on language pairs such as English-French where formality transfer has been more extensively studied, would provide an even better understanding of the capacity and limits of our proposed techniques.

The formality awareness-translation quality trandeoff that we observed in the back translation experiments might lead to interesting questions about whether the formality classes of Japanese and English map perfectly onto each other. We would like to perform further studies on whether more refined formality transfer techniques would improve formality transfer between Japanese and English, or whether a formality class mapping exists at all between Japanese and English.

\end{document}